# Nonapproximability Results for Partially Observable Markov Decision Processes

**Christopher Lusena**  LUSENA@CS.ENGR.UKY.EDU
*University of Kentucky*
*Dept. of Computer Science, Lexington KY 40506-0046*

**Judy Goldsmith**  GOLDSMIT@CS.ENGR.UKY.EDU
*University of Kentucky*
*Dept. of Computer Science, Lexington KY 40506-0046*

**Martin Mundhenk**  MUNDHENK@TI.UNI-TRIER.DE
*Universität Trier*
*FB IV – Informatik, D-54286 Trier, Germany*

## Abstract

We show that for several variations of partially observable Markov decision processes, polynomial-time algorithms for finding control policies are unlikely to or simply don't have guarantees of finding policies within a constant factor or a constant summand of optimal. Here "unlikely" means "unless some complexity classes collapse," where the collapses considered are P = NP, P = PSPACE, or P = EXP. Until or unless these collapses are shown to hold, any control-policy designer must choose between such performance guarantees and efficient computation.

## 1. Introduction

Life is uncertain; real-world applications of artificial intelligence contain many uncertainties. In this work, we show that uncertainty breeds uncertainty: In a controlled stochastic system with uncertainty (as modeled by a partially observable Markov decision process, for example), plans can be obtained efficiently or with quality guarantees, but rarely both.

Planning over stochastic domains with uncertainty is hard (in some cases PSPACE-hard or even undecidable, see Papadimitriou & Tsitsiklis, 1987; Madani, Hanks, & Condon, 1999). Given that it is hard to find an optimal plan or policy, it is natural to try to find one that is "good enough". In the best of all possible worlds, this would mean having an algorithm that is guaranteed to be fast and to produce a policy that is reasonably close to the optimal policy. Unfortunately, we show here that such an algorithm is unlikely or, in some cases, impossible. The implication for algorithm development is that developers should not waste time working toward both guarantees at once.

The particular mathematical models we concentrate on in this paper are *Markov decision processes (MDPs)* and *partially observable Markov decision processes (POMDPs)*. We consider both the straightforward representations of MDPs and POMDPs, and succinct representations, since the complexity of finding policies is measured not in terms of the size of the system, but in terms of the size of the *representation* of the system.

There has been a significant body of work on heuristics for succinctly represented MDPs (see Boutilier, Dean, & Hanks, 1999; Blythe, 1999 for surveys). Some of this work grows out of the engineering tradition (see, for example, Tsitsiklis and Van Roy's (1996) article on feature-based methods) which depends on empirical evidence to evaluate algorithms. While there are obvious drawbacks to this approach, our work argues that this may be the most appropriate way to verify the quality of an approximation algorithm, at least if one wants to do so in reasonable time.

The same problems that plague approximation algorithms for uncompressed representations carry over to the succinct representations, and the compression introduces additional complexity. For example, if there is no computable approximation of the optimal policy in the uncompressed case, then compression will not change this. However, it is easy to find the optimal policy for an infinite-horizon fully observable MDP (Bellman, 1957), yet EXP-hard (provably harder than polynomial time) to find approximately optimal policies (in time measured in the size of the input) if the input is represented succinctly (see Section 5).

Note that there are two interpretations to finding an approximation: finding a policy with value close to that of the optimal policy, or simply calculating a *value* that is close to the optimal value. If we can do the former and can evaluate policies, then we can certainly do the latter. Therefore, we sometimes show that the latter cannot be done, or cannot be done in time polynomial in the size of the input (unless something unlikely is true).

The complexity class PSPACE consists of those languages recognizable by a Turing machine that uses only $p(n)$ memory for some polynomial $p$, where $n$ is the size of the input. Because each time step uses at most one unit of memory, P $\subseteq$ PSPACE, though we do not know whether that is a proper inclusion or an equality. Because, given a limit on the amount of memory used, there are only exponentially many configurations of that memory possible with a fixed finite alphabet, PSPACE $\subseteq$ EXP. It is not known whether this is a proper inclusion or an equality either, although it is known that P $\neq$ EXP. Thus, a PSPACE-hardness result says that the problem is apparently not tractable, but an EXP-hardness result says that the problem is certainly not tractable.

Researchers also consider problems that are P-complete (under logspace or other highly restricted reductions). For example, the policy existence problem for infinite-horizon MDPs is P-complete (Papadimitriou & Tsitsiklis, 1987). This is useful information, because it is generally thought that P-complete problems are not susceptible to significant speed-up via parallelization. (For a more thorough discussion of P-completeness, see Greenlaw, Hoover, & Ruzzo, 1995.)

We also know that NP $\subseteq$ PSPACE, so P = PSPACE would imply P = NP. Thus, any argument or belief that P $\neq$ NP implies that P $\neq$ PSPACE. (For elaborations of this complexity theory primer, see any complexity theory text, such as Papadimitriou, 1994.)

In this paper, we show that there is a necessary trade-off between running time guarantees and performance guarantees for any general POMDP approximation algorithm — unless P = NP or P = PSPACE. (Table 1 gives an overview of our results.) Note that (assuming P $\neq$ NP or P $\neq$ PSPACE) this tells us that there is no algorithm that runs in time polynomial in the size of the representation of the POMDP that finds a policy that is close to optimal *for every instance*. It does not say that fast algorithms will produce far-from-optimal values for all POMDPs; there are many instances where the algorithms already in use or being developed will be both fast and close. We simply can't guarantee that the algorithms will *always* find a close-to-optimal policy quickly.

| policy | representation | horizon | problem | complexity |
|---|---|---|---|---|
| Partial observability | | | | |
| stationary | – | $n$ | $\varepsilon$-app. | not unless P=NP |
| stationary | – | $n$ | above-avg. value | not unless P=NP |
| time-dependent | – | $n$ | $\varepsilon$-app. | not unless P=NP |
| history-dependent | – | $n$ | $\varepsilon$-app. | not unless P=PSPACE |
| stationary | – | $\infty$ | $\varepsilon$-app. | not unless P=NP |
| time-dependent | – | $\infty$ | $\varepsilon$-app. | uncomputable (Madani et al., 1999) |
| Unobservability | | | | |
| time-dependent | – | $n$ | $\varepsilon$-app. | not unless P=NP |
| Full observability | | | | |
| stationary | – | $n$ | $k$-additive app. | P-hard |
| stationary | 2TBN | $n$ | $k$-additive app. | EXP-hard |

Table 1: Hardness for partially and fully observable MDPs

## 1.1 Heuristics and Approximations

The state of the art with respect to POMDP policy-finding algorithms is that there are three types of algorithms in use or under investigation: exact algorithms, approximations, and heuristics. Exact algorithms attempt to find exact solutions. In the finite-horizon cases, they run in worst-case time at least exponential in the size of the POMDP and the horizon (assuming a straightforward representation of the POMDP). In the infinite horizon, they do not necessarily halt, but can be stopped when the policy is within $\varepsilon$ of optimal (a checkable condition). Approximation algorithms construct approximations to what the exact algorithms find. (Examples of this include grid-based methods, Hauskrecht, 1997; Lovejoy, 1991; White, 1991.) Heuristics come in two flavors: those that construct or find actual policies that can be evaluated, and those that specify a means of choosing an action (for example, "most likely state"), which do not yield policies that can be evaluated using the standard, linear algebra-based methods.

The best current exact algorithm is incremental pruning (IP) with point-based improvement (Zhang, Lee, & Zhang, 1999). Littman's analysis of the witness algorithm (Littman, Dean, & Kaelbling, 1995; Cassandra, Kaelbling, & Littman, 1995) still applies: This algorithm requires exponential time in the worst case. The underlying theory of these algorithms (Witness, IP, etc.) for infinite-horizon cases depends on Bellman's and Sondik's work on value iteration for MDPs and POMDPs (Bellman, 1957; Sondik, 1971; Smallwood & Sondik, 1973).

The best known family of approximation algorithms is known as grid methods. The basic idea is to use a finite grid of points in the belief space (the space of all probability distributions over the states of the POMDP — this is the underlying space for the algorithms mentioned above) to define a policy. Once the grid points are chosen, all of these algorithms use value iteration on the points to obtain a policy for those belief states, then interpolate to the whole belief space. The difference in the algorithms lies in the choice of grid points. (An excellent survey appears in Hauskrecht, 1997.) These algorithms are called approximation algorithms because they approximate the process of value iteration, which the exact algorithms algorithms carry out exactly.

Heuristics that do not yield easily evaluated policies are surveyed in (Cassandra, 1998). These are often very easy to implement, and include techniques such as "most likely state" (choosing a state with the highest probability from the belief state, and acting as if the system were fully observable), and minimum entropy (choosing the action that gives the most information about the current state). Others depend on "voting," where several heuristics or options are combined.

There are heuristics based on finite histories or other uses of finite amounts of memory within the algorithm (Sondik, 1971; Platzman, 1977; Hansen, 1998a, 1998b; Lusena, Li, Sittinger, Wells, & Goldsmith, 1999; Meuleau, Kim, Kaelbling, & Cassandra, 1999; Meuleau, Peshkin, Kim, & Kaelbling, 1999; Peshkin, Meuleau, & Kaelbling, 1999; Hansen & Feng, 2000; Kim, Dean, & Meuleau, 2000). None of these comes with proofs of closeness, except for some of Hansen's work. For the rest, the trade-off has been made between fast searching through policy space and guarantees.

### 1.2 Structure of This Paper

In Section 2, we give formal definitions of MDPs and POMDPs and policies; two-phase temporal Bayes nets (2TBNs) are defined in Section 5. In Section 3, we define $\varepsilon$-approximations and additive approximations, and show a relationship between the two types of approximability for MDPs and POMDPs.

We separate the complexity results for finite-horizon policy approximation from those for infinite-horizon policies. Section 4 contains nonapproximability results for finite-horizon POMDP policies; Section 6 contains nonapproximability for infinite-horizon POMDP policies. Although it is relatively easy to find optimal MDP policies, we consider approximating MDP policies in Section 5, since the succinctly represented case, at least, is provably hard to approximate.

Some of the more technical proofs are included in appendices in order to make the body of the paper more readable. However, some proofs from other papers are sketched in the body of the paper in order to motivate both the results and the proofs newly presented here.

## 2. Definitions

Note that MDPs are in fact special cases of POMDPs. The complexity of finding and approximating optimal policies depends on the observability of the system, so our results are segregated by observability. However, one set of definitions suffices.

### 2.1 Partially Observable Markov Decision Processes

A partially observable Markov decision process (POMDP) describes a controlled stochastic system by its states and the consequences of actions on the system. It is denoted as a tuple $M = (\mathcal{S}, s_0, \mathcal{A}, \mathcal{O}, t, o, r)$, where

- $\mathcal{S}$, $\mathcal{A}$ and $\mathcal{O}$ are finite sets of *states*, *actions* and *observations*;

- $s_0 \in \mathcal{S}$ is the *initial state*;

- $t : \mathcal{S} \times \mathcal{A} \times \mathcal{S} \to [0,1]$ is the *state transition function*, where $t(s, a, s')$ is the probability that state $s'$ is reached from state $s$ on action $a$ (for every $s \in \mathcal{S}$ and $a \in \mathcal{A}$; either $\Sigma_{s' \in \mathcal{S}} t(s, a, s') = 1$, if action $a$ can be applied on state $s$, or $\Sigma_{s' \in \mathcal{S}} t(s, a, s') = 0$);

- $o : \mathcal{S} \to \mathcal{O}$ is the *observation function*, where $o(s)$ is the observation made in state $s$,[1]

- $r : \mathcal{S} \times \mathcal{A} \to \mathbf{Q}$ is the *reward function*, where $r(s, a)$ is the reward gained by taking action $a$ in state $s$.

If states and observations are identical, i.e. $\mathcal{O} = \mathcal{S}$ and $o$ is the identity function (or a bijection), then the MDP is called *fully observable*. Another special case is *unobservable MDPs*, where the set of observations contains only one element, i.e. in every state the same observation is made, and therefore the observation function is constant.

Normally, MDPs are represented by $\mathcal{S} \times \mathcal{S}$ tables, one for each action. However, we will also discuss more succinct representations: in particular, *two-phase temporal Bayes nets* (2TBNS). These will be defined in Section 5.

### 2.2 Policies and Performances

A policy describes how to act depending on observations. We distinguish three types of policies.

- A *stationary policy* $\pi_s$ (for $M$) is a function $\pi_s : \mathcal{O} \to \mathcal{A}$, mapping each observation to an action.

- A *time-dependent policy* $\pi_t$ is a function $\pi_t : \mathcal{O} \times \mathbf{N} \to \mathcal{A}$, mapping each pair ⟨observation, time⟩ to an action.

- A *history-dependent policy* $\pi_h$ is a function $\pi_h : \mathcal{O}^* \to \mathcal{A}$, mapping each finite sequence of observations to an action.

Notice that, for an unobservable MDP, a history-dependent policy is equivalent to a time-dependent one.

Recent algorithmic development has included consideration of finite memory policies as well (Hansen, 1998b, 1998a; Lusena, Li, Sittinger, Wells, & Goldsmith, 1999; Meuleau, Kim, Kaelbling, & Cassandra, 1999; Meuleau, Peshkin, Kim, & Kaelbling, 1999; Peshkin, Meuleau, & Kaelbling, 1999; Hansen & Feng, 2000; Kim, Dean, & Meuleau, 2000). These are policies that are allowed some finite amount of memory; sufficient allowances would enable such a policy to simulate a full history-dependent policy over a finite horizon, or perhaps a time-dependent policy, or to use less memory more judiciously. One variant of finite memory policies, which we call *free finite memory policies,* fixes the amount of memory *a priori*.

More formally, a free finite memory policy with the finite set $\mathcal{M}$ of memory states for POMDP $M = (\mathcal{S}, \mathcal{A}, \mathcal{O}, t, o, r)$ is a function $\pi_f : \mathcal{O} \times \mathcal{M} \to \mathcal{A} \times \mathcal{M}$, mapping each

---

1. Note that making observations probabilistically does not add any power to MDPs. Any probabilistically observable MDP can be turned into one with deterministic observations with only a polynomial increase in its size.

⟨observation, memory state⟩ pair to a pair ⟨action, memory state⟩. The set of memory states $\mathcal{M}$ can be seen as a finite "scratch" memory.

Free finite memory policies can also simulate stationary policies; all hardness results for stationary policies apply to free finite memory policies as well. Because one can consider a free finite memory policy to be a stationary policy over the state space $\mathcal{S} \times \mathcal{M}$, *all upper bounds (complexity class membership results) for stationary policies hold for free finite memory policies as well.* The advantages of free finite memory policies appear in the constants of the algorithms, and in special, probably large, subclasses of POMDPs, where a finite amount of memory suffices for an optimal policy. The maze instances such as McCallum's maze (McCallum, 1993; Littman, 1994) are such examples: McCallum's maze requires only 1 bit of memory to find an optimal policy.

Let $M = (\mathcal{S}, s_0, \mathcal{A}, \mathcal{O}, t, o, r)$ be a POMDP.

A *trajectory $\theta$ of length $m$ for $M$* is a sequence of states $\theta = \sigma_0, \sigma_1, \sigma_2, \ldots, \sigma_m$ ($m \geq 0$, $\sigma_i \in \mathcal{S}$) which starts with the initial state of $M$, i.e. $\sigma_0 = s_0$. We use $T_k(s)$ to denote the set of length-$k$ trajectories which end in state $s$.

The expected reward obtained in state $s$ after exactly $k$ steps under policy $\pi$ is the reward obtained in $s$ by taking the action specified by $\pi$, weighted by the probability that $s$ is actually reached after $k$ steps,

- $r(s, k, \pi) = r(s, \pi(o(s))) \cdot \sum_{(\sigma_0, \ldots, \sigma_k) \in T_k(s)} \prod_{i=1}^{k} t(\sigma_{i-1}, \pi(o(\sigma_{i-1})), \sigma_i)$, if $\pi$ is a stationary policy,

- $r(s, k, \pi) = r(s, \pi(o(s), k)) \cdot \sum_{(\sigma_0, \ldots, \sigma_k) \in T_k(s)} \prod_{i=1}^{k} t(\sigma_{i-1}, \pi(o(\sigma_{i-1}), i-1), \sigma_i)$, if $\pi$ is a time-dependent policy, and

- $r(s, k, \pi) = \sum_{(\sigma_0, \ldots, \sigma_k) \in T_k(s)} r(s, \pi(o(\sigma_0) \cdots o(\sigma_k))) \cdot \prod_{i=1}^{k} t(\sigma_{i-1}, \pi(o(\sigma_0) \cdots o(\sigma_{i-1})), \sigma_i)$, if $\pi$ is a history-dependent policy.

A POMDP may behave differently under optimal policies for each type of policy. The quality of a policy is determined by its *performance*, i.e. by the expected rewards accrued by it. We distinguish between different performance metrics for POMDPs that run for a finite number of steps and those that run indefinitely.

- The *finite-horizon performance of a policy* $\pi$ for POMDP $M$ is the expected sum of rewards received during the first $|M|$ steps by following the policy $\pi$, i.e., $perf_f(M, \pi) = \sum_{i=0}^{|M|-1} \sum_{s \in \mathcal{S}} r(s, i, \pi)$. (Other work assumes that the horizon is $poly(|M|)$, instead of $|M|$. This does not change the complexity of any of our problems.)

- The infinite-horizon *total discounted performance* gives rewards obtained earlier in the process a higher weight than those obtained later. For $0 < \beta < 1$, the total $\beta$-discounted reward is defined as $perf_{td}^{\beta}(M, \pi) = \sum_{i=0}^{\infty} \sum_{s \in \mathcal{S}} \beta^i \cdot r(s, i, \pi)$.

- The infinite-horizon *average performance* is the limit of all rewards obtained within $n$ steps divided by $n$, for $n$ going to infinity:[2] $perf_{av}(M, \pi) = \lim_{n \to \infty} \frac{1}{n} perf_f(M, n, \pi)$.

---
2. If this limit is not defined, the performance is defined as a lim inf.

NONAPPROXIMABILITY RESULTS FOR POMDPSLet *perf* be any of these performance metrics, and let $\alpha$ be any policy type, either stationary, time-dependent, or history-dependent. The $\alpha$-*value* $\text{val}_\alpha(M)$ of $M$ (under the metric chosen) is the maximal performance of any policy $\pi$ of type $\alpha$ for $M$, i.e. $\text{val}_\alpha(M) = \max_{\pi \in \Pi_\alpha} perf(M, \pi)$, where $\Pi_\alpha$ is the set of all $\alpha$ policies.

For simplicity, we assume that the size $|M|$ of a POMDP $M$ is determined by the size $n$ of its state space. We assume that there are no more actions than states, and that each state transition probability is given as a binary fraction with $n$ bits and each reward is an integer of at most $n$ bits. This is no real restriction, since adding unreachable "dummy" states allows one to use more bits for transition probabilities and rewards. Also, it is straightforward to transform a POMDP $M$ with non-integer rewards to $M'$ with integer rewards such that $\text{val}_\alpha(M, k) = c \cdot \text{val}_\alpha(M', k)$ for some constant $c$ depending only on $(M, k)$ and not on $\text{val}_\alpha(M, k)$.

We consider problem instances that are represented in a straightforward way. A POMDP with $n$ states is represented by a set of $n \times n$ tables for the transition function (one table for each action) and a similar table for the reward function and for the observation function. We assume that the number of actions and the number of bits needed to store each transition probability or reward does not exceed $n$, so such a representation requires $O(n^4)$ bits. (This can be modified to allow $n^k$ bits without changing the complexity results.) In the same way, stationary policies can be encoded as lists with $n$ entries, and time-dependent policies for horizon $n$ as $n \times n$ tables.

For each type of POMDP, each type of policy, and each type of performance metric the *value problem* is,

**given** a POMDP, a performance metric (finite-horizon, total discounted, or average performance), and a policy type (stationary, time-dependent, or history-dependent),

**calculate** the value of the best policy of the specified type under the given performance metric.

The *policy existence problem* is,

**given** a POMDP, a performance metric, and a policy type,

**decide** whether the value of the best policy of the specified type under the given performance metric is greater 0.

## 3. Approximability

In previous work (Papadimitriou & Tsitsiklis, 1986, 1987; Mundhenk, Goldsmith, & Allender, 1997; Mundhenk, Goldsmith, Lusena, & Allender, 2000; Madani et al., 1999), it was shown that the policy existence problem is computationally intractable for most variations of POMDPs, or even undecidable for some infinite-horizon cases. For example, we showed that the stationary policy existence problems for POMDPs with or without negative rewards are NP-complete. Computing an optimal policy is at least as hard as deciding the existence problem. Instead of asking for an optimal policy, we might wish to compute a policy that is guaranteed to have a value that is at least a large fraction of the optimal value.

A polynomial-time algorithm computing such a nearly optimal policy is called an $\varepsilon$-approximation (for $0 \leq \varepsilon < 1$), where $\varepsilon$ indicates the quality of the approximation in the following way. Let $A$ be a polynomial-time algorithm which for every POMDP $M$ computes an $\alpha$-policy $A(M)$. Notice that $perf(M, A(M)) \leq \text{val}_\alpha(M)$ for every $M$. The algorithm $A$ is called an $\varepsilon$-*approximation* if for every POMDP $M$,

$$\left| \frac{\text{val}_\alpha(M) - perf(M, A(M))}{\text{val}_\alpha(M)} \right| \leq \varepsilon \ .$$

(See, e.g., Papadimitriou, 1994 for more detailed definitions.) Approximability distinguishes NP-complete problems: There are problems which are $\varepsilon$-approximable for all $\varepsilon$, for certain $\varepsilon$, or for no $\varepsilon$ (unless P = NP). Note that this definition of $\varepsilon$-approximation requires that $\text{val}_\alpha(M) \geq 0$. If a policy with positive performance exists, than every approximation algorithm yields such a policy, because a policy with performance 0 or smaller cannot approximate a policy with positive performance. Hence, any approximation straightforwardly solves the decision problem.

An approximation scheme yields an $\varepsilon$-approximation for arbitrary $\varepsilon > 0$. If there is a polynomial-time algorithm that on input POMDP $M$ and $\varepsilon$ outputs an $\varepsilon$-approximation of the value, in time polynomial in the size of $M$ then we say the problem has a *Polynomial-Time Approximation Scheme (PTAS)*. If the algorithm runs in time polynomial in the size of $M$ and $\frac{1}{\varepsilon}$, the scheme is a *Fully Polynomial-Time Approximation Scheme (FPTAS)*. All of the PTASs constructed here are FPTASs; we state the theorems in terms of PTASs because that gives stronger results in some cases, and because we do not explicitly analyze the complexity in terms of $\frac{1}{\varepsilon}$.

If there is a polynomial-time algorithm that outputs an approximation, $v$, to the value $\mu$ of $M$ ($\mu = \text{val}_\alpha(M)$) with $\mu \geq v \geq \mu - k$, then we say that the problem has a *k-additive approximation algorithm*.

In the context of POMDPs, existence of a $k$-additive approximation algorithm and a PTAS are often equivalent. This might seem surprising to readers who are more familiar with reward criteria that have fixed upper and lower bounds on the performance of a solution, for example, the probability of reaching a goal state. In these cases, the fixed bounds on performance will give different results. However, we are addressing the case where there is no *a priori* upper bound on the performance of policies, even though there are computable upper bounds on the performance of a policy for each *instance*.

**Theorem 3.1** *For POMDPs with non-negative rewards and flat representations under finite-horizon total or total discounted, and infinite-horizon total discounted reward metrics, if there exists a k-additive approximation, then we can determine in polynomial time whether there is a policy with performance greater than 0.*

**Proof** The theorem follows from two facts: (1) given a POMDP $M$ with value $\mu$, we can construct another POMDP $\theta M$ with value $\theta \cdot \mu$ just by multiplying all rewards in the former POMDP by $\theta$; (2) under these reward metrics we can find a lower bound on $\mu$ if it is not 0.

The computation of the lower bound, $\delta$, on the value of $\mu$ depends on the reward metric. Because there are no negative rewards, in order for the expected reward to be positive in the finite-horizon case, an action with positive reward must be taken with nonzero probability by the last step. Consider only reachable states of the POMDP, and let $\nu$ be the lowest nonzero transition probability to one of these states, $h$ the horizon, and $\zeta$ the smallest nonzero reward, and set $\delta = \nu^h \zeta$. Then $\nu^h$ is a lower bound on the probability of actually reaching any particular state after $h$ steps (if this probability is nonzero), in particular a state with reward $\zeta$. If the reward metric is discounted, then let $\delta = (\gamma\nu)^h \zeta$, where $\gamma \in (0, 1]$ is the discount factor.

Now consider the infinite-horizon under a stationary policy. This induces a Markov process, and the policy has nonzero reward if there is a nonzero probability path to a reward node, i.e., a state from which there is a positive-reward action possible. This is true if and only if there is a nonzero-probability *simple* path (visiting each node at most once) to a reward node. Such a path accrues reward at least $\delta = (\gamma\nu)^{|S|}\zeta$ for stationary policies.

Since stationary policies have values bounded by the time-dependent and history-dependent values for infinite-horizon POMDPs, this lower bound for the stationary value of the POMDP is also a lower bound for other policies.

Finally, note that if the value of a POMDP with non-negative rewards is 0, then a $k$-additive approximation cannot return a positive value. To determine whether there is a policy with reward greater than 0 for a given POMDP, compute $\delta$ and then set $\theta$ such that $\theta\delta - k > 0$, i.e., $\theta > \frac{k}{\delta}$, and run the $k$-additive approximation algorithm on $\theta M$. The POMDP has positive value if and only if the approximation returns a positive value. □

Note that this does not contradict the undecidability result of Madani et al. (1999). The problem that they proved undecidable is whether a POMDP with nonpositive rewards has a history-dependent or time-dependent value of 0. We're asking whether it has value $> 0$ in the non-negative reward case; answering this question (even if we multiply the rewards by $-1$) does not answer their question.

**Corollary 3.2** *For POMDPs with flat representations and non-negative rewards, the POMDP value problem under finite-horizon or infinite-horizon total discounted reward is k-additive approximable if and only if there exists a PTAS for that POMDP value problem.*

Note that the corollary depends only on Facts (1) and (2) from the proof of Theorem 3.1. Thus, any optimization problem with those properties will have a $k$-additive approximation if and only if it has a PTAS.

**Proof** Let $\mu = \text{val}_\alpha(M)$, and let $A$ be a polynomial-time $k$-additive approximation algorithm. First, the PTAS computes $\delta$ as in Theorem 3.1 and checks whether $\mu = 0$. If so, it outputs 0. Otherwise, given $\varepsilon$, it chooses $\theta$ such that $\theta > \frac{k}{\varepsilon\delta} \geq \frac{k}{\varepsilon\mu}$, and thus $\theta\mu - k > (1-\varepsilon)\theta\mu$ holds. Let $v = A(\theta M)$. (Note that $A(M)$ is the approximation to the value of $M$ found by running algorithm $A$.) Then $v$ is an $\varepsilon$-approximation to $\theta\mu$, so $\frac{v}{\theta}$ is an $\varepsilon$-approximation to $\mu$.

Suppose, instead, that we have a PTAS for optimal policies for this problem. Let $A(M, \varepsilon)$ be an algorithm that demonstrates this. Let $\mu = \text{val}_\alpha(M)$, and $A(M, 0.5) = v$. Thus $\mu \geq v \geq \frac{\mu}{2}$. If $\mu = 0$ then $v = 0$ and we can stop. Else we choose an $\varepsilon$ such that $(1-\varepsilon)\mu \geq \mu - k$, giving $\varepsilon \leq \frac{k}{\mu}$. Since $\frac{k}{2v} \leq \frac{k}{\mu}$, and $\frac{k}{2v}$ is polynomial size and is polynomial-time computable in $|M|$ (since $v$ is the output of $A(M, 0.5)$), we can choose $\varepsilon < \frac{k}{2v}$, and run $A(M, \varepsilon)$. This gives a $k$-additive approximation. □

A problem that is not $\varepsilon$-approximable for some $\varepsilon$ cannot have a PTAS. Therefore, any multiplicative nonapproximability result yields an additive nonapproximability result. However, an additive nonapproximability result only shows that there is no PTAS, although there might be an $\varepsilon$-approximation for some fixed $\varepsilon$.

## 4. Non-Approximability for Finite-Horizon POMDPs

This section focuses on finite-horizon policies. Because that is consistent throughout the section, we do not explicitly mention it in each theorem. However, as Section 6 shows, there are significant computational differences between finite- and infinite-horizon calculations.

The policy existence problem for POMDPs with negative and non-negative rewards is not suited for $\varepsilon$-approximation. If a policy with positive performance exists, then every approximation algorithm yields such a policy, because a policy with performance 0 or smaller cannot approximate a policy with positive performance. Hence, the decision problem is straightforwardly solved by any $\varepsilon$-approximation. Therefore, we concentrate on POMDPs with non-negative rewards. Results for POMDPs with unrestricted rewards are stated as corollaries. Consider an $\varepsilon$-approximation algorithm $A$ that, on input a POMDP $M$ with non-negative rewards, outputs a policy $\pi_\alpha^M$ of type $\alpha$. Then it holds that

$$perf(M, \pi_\alpha^M) \geq (1 - \varepsilon) \cdot \mathrm{val}_\alpha(M).$$

We first consider the question of whether an optimal stationary policy can be $\varepsilon$-approximated for POMDPs with non-negative rewards. It is known (Littman, 1994; Mundhenk et al., 2000) that the related decision problem is NP-complete. We include a sketch of that proof here, since later proofs build on it. The formal details can be found in Appendix A.

**Theorem 4.1** *(Littman, 1994; Mundhenk et al., 2000) The stationary policy existence problem for POMDPs with non-negative rewards is* NP*-complete.*

**Proof** Membership in NP is straightforward, because a policy can be guessed and evaluated in polynomial time. To show NP-hardness, we reduce the NP-complete satisfiability problem 3Sat to it. Let $\phi(x_1, \ldots, x_n)$ be such a formula with variables $x_1, \ldots, x_n$ and clauses $C_1, \ldots, C_m$, where clause $C_j = (l_{v(1,j)} \vee l_{v(2,j)} \vee l_{v(3,j)})$ for $l_i \in \{x_i, \neg x_i\}$. We say that variable $x_i$ appears in $C_j$ with signum 0 (resp. 1) if $\neg x_i$ (resp. $x_i$) is a literal in $C_j$. Without loss of generality, we assume that every variable appears at most once in each clause. The idea is to construct a POMDP $M(\phi)$ having one state for each appearance of a variable in a clause. The set of observations is the set of variables. Each action corresponds to an assignment of a value to a variable. The transition function is deterministic. The process starts with the first variable in the first clause. If the action chosen in a certain state satisfies the corresponding literal, the process proceeds to the first variable of the next clause, or with reward 1 to a final sink state $T$ if all clauses were considered. If the action does not satisfy the literal, the process proceeds to the next variable of the clause, or with reward 0 to a sink state $F$. A sink state will never be left. The partition of the state space into observation classes guarantees that the same assignment is made for every appearance of the same variable. Therefore, the value of $M(\phi)$ equals 1 iff $\phi$ is satisfiable. The formal reduction is in Appendix A. □

Note that all policies have expected reward of either 1 or 0. Immediately we get the nonapproximability result for POMDPs, even if all trajectories have non-negative performance.

**Theorem 4.2** *Let $0 \leq \varepsilon < 1$. An optimal stationary policy for POMDPs with non-negative rewards is $\varepsilon$-approximable if and only if* P = NP.

**Proof** The stationary value of a POMDP can be calculated in polynomial time by a binary search using an oracle for the stationary policy existence problem for POMDPs. The number of bits to be calculated is polynomial in the size of $M$. Knowing the value, we can try to fix an action for an observation. If the modified POMDP still achieves the value calculated before, we can continue with the next observation, until a stationary policy is found which has the optimal performance. This algorithm runs in polynomial time with an oracle solving the stationary policy existence problem for POMDPs. Since the oracle is in NP, by Theorem 4.1, the algorithm runs in polynomial time if P = NP.

Now, assume that $A$ is a polynomial-time algorithm that $\varepsilon$-approximates the optimal stationary policy for some $\varepsilon$ with $0 \leq \varepsilon < 1$. We show that this implies that P = NP by showing how to solve the NP-complete problem 3SAT. As in the proof of Theorem 4.1, given an instance $\phi$ of 3SAT, we construct a POMDP $M(\phi)$. The only change to the reward function of the POMDP constructed in the proof of Theorem 4.1 is to make it a POMDP with positive performances. Now reward 1 is obtained if state $F$ is reached, and reward $\lceil \frac{2}{1-\varepsilon} \rceil$ is obtained if state $T$ is reached. Hence $\phi$ is satisfiable if and only if $M(\phi)$ has value $\lceil \frac{2}{1-\varepsilon} \rceil$.

Assume that policy $\pi$ is the output of the $\varepsilon$-approximation algorithm $A$. If $\phi$ is satisfiable, then $\mathit{perf}(M(\phi), \pi) \geq (1-\varepsilon) \cdot \frac{2}{1-\varepsilon} = 2 > 1$. Because the performance of every policy for $M(\phi)$ is either 1 if $\phi$ is not satisfiable, or $\lceil \frac{2}{1-\varepsilon} \rceil$ if $\phi$ is satisfiable, it follows that $\pi$ has performance $> 1$ if and only if $\phi$ is satisfiable. So, in order to decide $\phi \in$ 3SAT, we can construct $M(\phi)$, run the approximation algorithm $A$ on it, take its output $\pi$ and calculate $\mathit{perf}(M(\phi), \pi)$. That output shows whether $\phi$ is in 3SAT. All these steps are polynomial-time bounded computations. It follows that 3SAT is in P, and hence P = NP. $\square$

Of course, the same nonapproximability result holds for POMDPs with positive and negative rewards.

**Corollary 4.3** *Let $0 \leq \varepsilon < 1$. Any optimal stationary policy for POMDPs is $\varepsilon$-approximable if and only if* P = NP.

Using the same proof technique as above, we can show that the value is nonapproximable, too.

**Corollary 4.4** *Let $0 \leq \varepsilon < 1$. The stationary value for POMDPs is $\varepsilon$-approximable if and only if* P = NP.

A similar argument can be used to show that a policy with performance at least the average of all performances for a POMDP cannot be computed in polynomial time, unless P = NP. Note that in the proof of Theorem 4.1, the only performance greater than or equal to the average of all performances is that of an optimal policy.

**Corollary 4.5** *The following are equivalent.*

1. *There exists a polynomial-time algorithm that for a given POMDP M computes a stationary policy under which M has performance greater than or equal to the average stationary performance of M.*

2. *P = NP.*

Thus, even calculating a policy whose performance is above average is likely to be infeasible.

We now turn to time-dependent policies. The time-dependent policy existence problem for POMDPs is known to be NP-complete, as is the stationary one.

**Theorem 4.6** *(Mundhenk et al., 2000) The time-dependent policy existence problem for unobservable MDPs is* NP-*complete.*

Papadimitriou and Tsitsiklis (1987) proved a theorem similar to Theorem 4.6. Their MDPs had only non-positive rewards, and their formulation of the decision problem was whether there is a policy with performance 0. The proof by Mundhenk et al., 2000, like theirs, uses a reduction from 3SAT. We modify this reduction to show that an optimal time-dependent policy is hard to approximate even for unobservable MDPs.

**Theorem 4.7** *Let $0 \leq \varepsilon < 1$. Any optimal time-dependent policy for unobservable MDPs with non-negative rewards is $\varepsilon$-approximable if and only if* P = NP.

**Proof** We give a reduction from 3SAT with the following properties. For a formula $\phi$ with $m$ clauses we show how to construct an unobservable MDP $M_\varepsilon(\phi)$ with value 1 if $\phi$ is satisfiable, and with value $< (1 - \varepsilon)$ if $\phi$ is not satisfiable. Therefore, an $\varepsilon$-approximation could be used to distinguish between satisfiable and unsatisfiable formulas in polynomial time.

For formula $\phi$, we first show how to construct an unobservable $M(\phi)$ from which $M_\varepsilon(\phi)$ will be constructed. (The formal presentation appears in Appendix B.) $M(\phi)$ simulates the following strategy. At the first step, one of the $m$ clauses is chosen uniformly at random with probability $\frac{1}{m}$. At step $i + 1$, the assignment of variable $i$ is determined. Because the process is unobservable, it is guaranteed that each variable gets the same assignment in all clauses, because its value is determined in the same step. If a clause is satisfied by this assignment, a final state will be reached. If not, an error state will be reached.

Now, construct $M_\varepsilon(\phi)$ from $m^2$ copies $M_1, \ldots, M_{m^2}$ of $M_\phi$, such that the initial state of $M_\varepsilon(\phi)$ is the initial state of $M_1$, the initial state of $M_{i+1}$ is the final state $T$ of $M_i$, and reward 1 is gained if the final state of $M_{m^2}$ is reached. The error states of all the $M_i$s are identified as a unique sink state $F$.

To illustrate the construction, in Figure 1 we give an example POMDP consisting of a chain of 4 copies of $M(\phi)$ obtained for the formula $\phi = (\neg x_1 \lor x_3 \lor x_4) \land (x_1 \lor \neg x_2 \lor x_4)$. The dashed arrows indicate a transition with probability $\frac{1}{2}$. The dotted (resp. solid) arrows are probability 1 transitions on action 0 (resp. 1). The actions correspond to assignments to the variables.

If $\phi$ is satisfiable, then a time-dependent policy simulating $m^2$ repetitions of any satisfying assignment has performance 1. If $\phi$ is not satisfiable, then under any assignment at least one of the $m$ clauses of $\phi$ is not satisfied. Hence, the probability that under any time-dependent policy the final state $T$ of $M(\phi)$ is reached is at most $1 - \frac{1}{m}$. Consequently, the probability that the final state of $M_\varepsilon(\phi)$ is reached is at most $(1 - \frac{1}{m})^{m^2} \leq e^{-m}$. This probability equals the expected reward. Since for large enough $m$ it holds that $e^{-m} < (1 - \varepsilon)$, the theorem follows. □

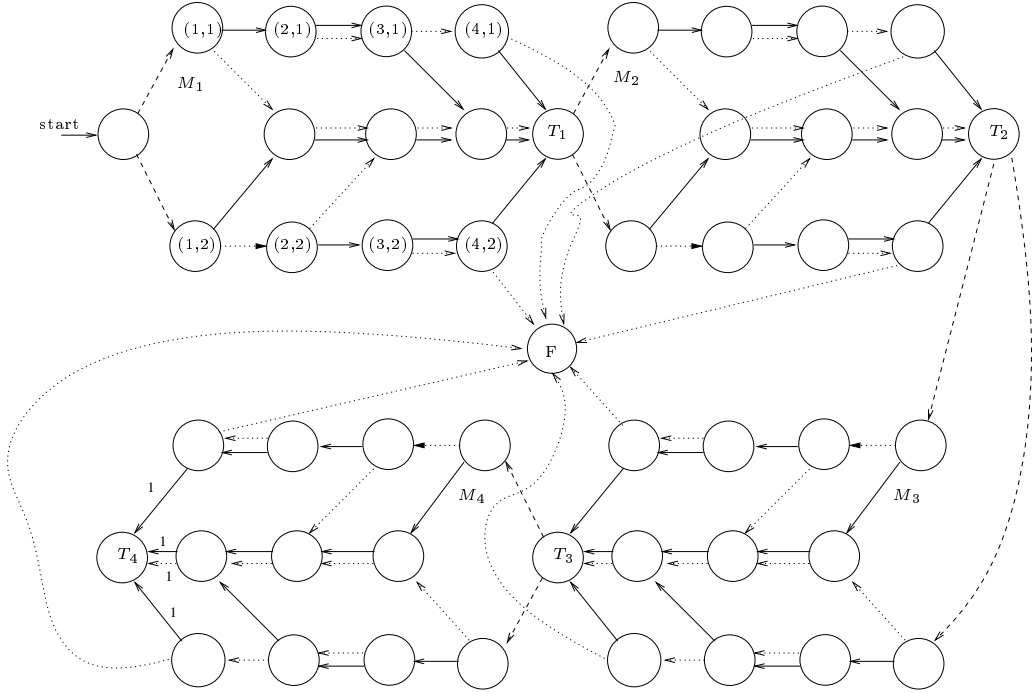

Figure 1: An example unobservable MDP for $\phi = (\neg x_1 \vee x_3 \vee x_4) \wedge (x_1 \vee \neg x_2 \vee x_4)$

Note that the time-dependent policy existence problem for POMDPs with non-negative rewards is NL-complete (Mundhenk et al., 2000). The class NL consists of those languages recognizable by nondeterministic Turing machines that use a read-only input tape and additional read-write tapes with $O(\log n)$ tape cells. It is known that NL $\subseteq$ P and that NL is properly contained in PSPACE. Unlike the case of stationary policies, approximability of time-dependent policies is harder than the policy existence problem (unless NL = NP).

Unobservability is a special case of partial observability. Hence, we get the same nonapproximability result for POMDPs, even for unrestricted rewards.

**Corollary 4.8** *Let $0 \leq \varepsilon < 1$. Any optimal time-dependent policy for POMDPs is $\varepsilon$-approximable if and only if* P = NP.

**Corollary 4.9** *Let $0 \leq \varepsilon < 1$. The time-dependent value of POMDPs is $\varepsilon$-approximable if and only if* P = NP.

Note that the proof of Theorem 4.7 assumed a total expected reward criterion. The discounted reward criterion is also useful in the finite horizon. To show the result for a discounted reward criterion, we only need to change the reward in the proof of Theorem 4.7 as follows: Multiply the final reward by $\beta^{-m^2(n+1)}$, where $\beta$ is the discount factor, $m$ the number of clauses, and $n$ the number of variables of the formula $\phi$.

Papadimitriou and Tsitsiklis (1987) proved that a problem very similar to history-dependent policy existence is PSPACE-complete.

**Theorem 4.10** *(Papadimitriou & Tsitsiklis, 1987; Mundhenk et al., 2000) The history-dependent policy existence problem for POMDPs is PSPACE-complete.*

To describe a horizon $n$ history-dependent policy for a POMDP with $c$ observations explicitly takes space $\sum_{i=1}^{n} c^i$. (We do not address the case of succinctly represented policies for POMDPs here. For an analysis of their complexity, see Mundhenk, 2000a.) If $c > 1$, this is exponential space. Therefore, we cannot expect that a polynomial-time algorithm outputs a history-dependent policy, and we restrict consideration to polynomial-time algorithms that approximate the history-dependent *value* — the optimal performance under any history-dependent policy — of a POMDP. Burago, de Rougemont, and Slissenko (1996) considered the class of POMDPs with a bound of $q$ on the number of states corresponding to an observation, where the rewards corresponded to the probability of reaching a fixed set of goal states (and thus were bounded by 1). They showed that for any fixed $q$, the optimal history-dependent policies for POMDPs in this class can be approximated to within an additive constant $k$. We showed in Proposition 3.2 that POMDP history-dependent discounted or total-reward value problems that can be approximated to within an additive constant $k$ have polynomial-time approximation schemes (Proposition 3.2), *as long as there are no a priori bounds on either the number of states per observation or the rewards.*

Notice, however, that Theorem 4.11 does not give us information about the classes of POMDPs that Burago et al. (1996) considered: Because of the restrictions associated with the parameter $q$, our hardness results do not contradict their result.

Finally, we show that the history-dependent value of POMDPs with non-negative rewards is not $\varepsilon$-approximable under total expected or discounted rewards, unless P = PSPACE. Consequently, the value has no PTAS or $k$-additive approximation under the same assumption.

The history-dependent policy existence problem for POMDPs with non-negative rewards is NL-complete (Mundhenk et al., 2000). Hence, because NL is a proper subclass of PSPACE, approximability of the history-dependent value is *strictly* harder than the policy existence problem.

**Theorem 4.11** *Let $0 \leq \varepsilon < 1$. The history-dependent value of POMDPs with non-negative rewards is $\varepsilon$-approximable if and only if P = PSPACE.*

**Proof** The history-dependent value of a POMDP $M$ can be calculated using binary search over the history-dependent policy existence problem. The number of bits to be calculated is polynomial in the size of $M$. Therefore, by Theorem 4.10, this calculation can be performed in polynomial time using a PSPACE oracle. If P = PSPACE, it follows that the history-dependent value of a POMDP $M$ can be *exactly* calculated in polynomial time.

The set QSAT of true quantified Boolean formulae is one of the standard PSPACE complete sets. To conclude P = PSPACE from an $\varepsilon$-approximation of the history-dependent value problem, we use a transformation of instances of QSAT to POMDPs similar to the proof of Theorem 4.10 in (Mundhenk, 2000b).

The set QSAT can be interpreted as a two-player game: Player 1 sets the existentially quantified variables, and player 2 sets the universally quantified variables. Player 1 wins if the alternating choices determine a satisfying assignment to the formula, and player 2 wins if the determined assignment is not satisfying. A formula is in QSAT if and only if player 1 has a winning strategy. This means player 1 has a response to every choice of player 2, so that in the end the formula will be satisfied.

The version where player 2 makes random choices and player 1's goal is to win with probability $> \frac{1}{2}$ corresponds to SSAT (*stochastic satisfiability*), which is also PSPACE complete. The instances of SSAT are formulas which are quantified alternatingly with existential quantifiers $\exists$ and random quantifiers $R$. The meaning of the random quantifier $R$ is that an assignment to the respective variable is chosen uniformly at random from $\{0, 1\}$. A stochastic Boolean formula
$$\Phi = \exists x_1 R x_2 \exists x_3 R x_4 \ldots \phi$$
is in SSAT if and only if

there exists $b_1$ for random $b_2$ exists $b_3$ for random $b_4 \ldots Prob[\phi(b_1, \ldots, b_n) \text{ is true}] > \frac{1}{2}$.

If $\Phi$ has $r$ random quantifiers, then the strategy of player 1 determines a set of $2^r$ assignments to $\phi$. The term "$Prob[\phi(b_1, \ldots, b_n) \text{ is true}] > \frac{1}{2}$" means that more than $2^{r-1}$ of these $2^r$ assignments satisfy $\phi$.

From the proof of IP = PSPACE by Shamir (1992) it follows that for every PSPACE set $A$ and constant $c \geq 1$ there is a polynomial-time reduction $f$ from $A$ to SSAT such that for every instance $x$ and formula $f(x) = \exists x_1 R x_2 \ldots \phi_x$ the following holds.

- If $x \in A$, then $\exists b_1$ for random $b_2 \ldots Prob[\phi_x(b_1, \ldots, b_n) \text{ is true}] > (1 - 2^{-c})$, and

- if $x \notin A$, then $\forall b_1$ for random $b_2 \ldots Prob[\phi_x(b_1, \ldots, b_n) \text{ is true}] < 2^{-c}$.

This means that player 1 either has a strategy under which she wins with very high probability, or the probability of winning (under any strategy) is very small. We show how to transform a stochastic Boolean formula $\Phi$ into a POMDP with a large history-dependent value if player 1 has a winning strategy, and a much smaller value if player 2 wins.

For an instance $\Phi = \exists x_1 R x_2 \ldots \phi$ of SSAT, where $\Phi$ is a formula with $n$ variables $x_1, \ldots, x_n$, we construct a POMDP $M(\Phi)$ as follows. The role of player 1 is taken by the controller of the process. A strategy of player 1 determines a policy of the controller, and vice versa. Player 2 appears as probabilistic transitions in the process. The process $M(\Phi)$ has three stages. The first stage consists of one step. The process chooses uniformly at random one of the variables and an assignment to it, and stores the variable and the assignment. More formally, from the initial state $s_0$, one of the states "$x_i = b$" ($1 \leq i \leq n$, $b \in \{0, 1\}$) is reached, each with probability $1/(2n)$. It is not observable which variable assignment was stored by the process. However, whenever that variable appears later, the process checks that the initially fixed assignment is chosen again. If the policy gives a different assignment during the second stage, the process halts with reward 0. (There is a deterministic transition to a final state which we refer to as $s_{end}$, or less formally, the *dead end* state.) If such an inconsistency occurs during the third stage, the process halts with reward 0 and notices that the policy *cheats*. (There is a deterministic transition to a sink state which we refer to as $s_{cheat}$, or less formally as *the penalty box* because the player sent there cannot re-enter the game later.) If eventually the whole formula is passed, either reward 2 or reward 0 is obtained dependent on whether the formula was satisfied or not. The first stage is sketched in Figure 2. In this and the following figures, dashed arrows represent random transitions (all of equal probability, irregardless of the action chosen), solid arrows represent deterministic transitions corresponding to the action 1 (True), and dotted arrows represent deterministic transitions corresponding to the action 0 (False).

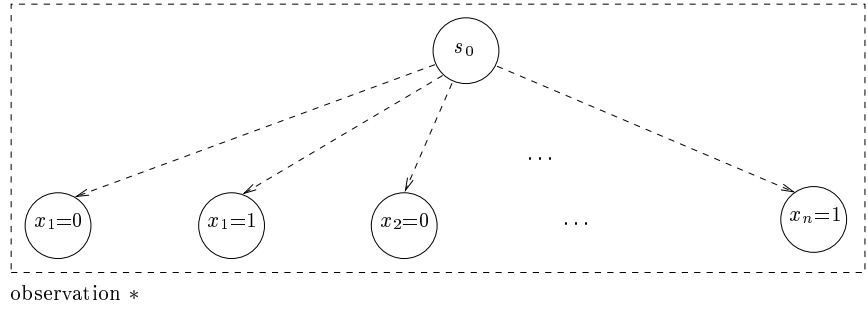

Figure 2: The first stage of $M(\Phi)$.

The second stage starts in each of the states "$x_i = b$" and has $n$ steps, during which an assignment to the variables $x_1, x_2, \ldots, x_n$ is fixed. Let $A_{c,b}$ denote the part of the process' second stage during which it is assumed that value $b$ is assigned to variable $x_c$. If a variable $x_i$ is existentially quantified, then the assignment is the action in $\{0, 1\}$ chosen by the policy. If a variable $x_i$ is randomly quantified, then the assignment is chosen uniformly at random by the process, independent of the action of the policy. In the second stage, it is observable which assignment was made to every variable. If the variable assignment from the first stage does not coincide with the assignment made to that variable during the second stage, the trajectory on which that happens ends in the dead end state that yields reward 0. Let $r$ be the number of random quantifiers of $\Phi$. Every strategy of player 1 determines $2^r$ assignments. Every assignment $(x_1 = b_1, \ldots, x_n = b_n)$ induces $2n$ trajectories: $n$ have the form

$$s_0, x_i{=}b_i, [x_1, b_1], \ldots, [x_i, b_i], \ldots, [x_n, b_n]$$

(for $i = 1, 2, \ldots, n$) that pass stage 2 without reaching the dead end state and continue with the first state of the third stage, and $n$ that dead-end in stage 2. The latter $n$ trajectories that do not reach stage 3 are of the form

$$s_0, x_i{=}b_i, [x_1, b_1], \ldots, [x_i, 1 - b_i], s_{end}$$

(for $i = 1, 2, \ldots, n$). Accordingly, $M(\Phi)$ has $n \cdot 2^r$ trajectories that reach the third stage. The structure of stage 2 is sketched for $x_3 = 0$ in Figure 3.

The third stage checks whether $\phi$ is satisfied by that trajectory's assignment. The process passes sequentially through the whole formula checking each literal in each clause for an assignment to the respective variable.[3] The case of a *cheating policy*, i.e., one that answers during the third stage with another assignment than fixed during the second stage, must be "punished". Whenever the variable corresponding to the initial, stored assignment appears the process checks that the stored assignment is consistent with the current assignment. If eventually the whole formula passes the checking, either reward 2 or reward 0 is obtained, depending on whether the formula was satisfied and the policy was not cheating, or not. Let $C_{c,b}$ be that instance of the third stage where it is checked whether $x_c$ always gets assignment $b$. It is essentially the same deterministic process as defined in the proof of Theorem 4.1, but whenever an assignment to a literal containing $x_c$ is asked for, if $x_c$ does not get assignment $b$ the process goes to state $s_{cheat}$. Otherwise, the process goes to state $s_{end}$. If the assignment chosen by the policy satisfied the formula, reward 2 is obtained; otherwise the reward is 0.

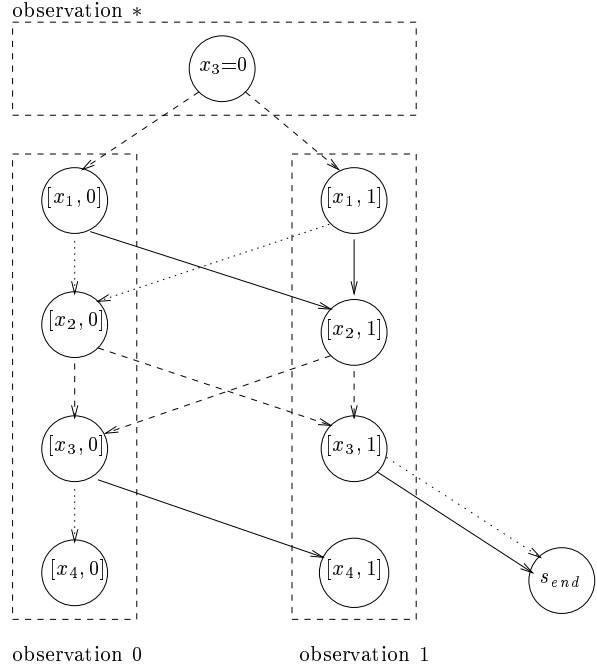

Figure 3: The second stage of $M(\Phi)$: $A_{3,0}$ for the quantifier prefix $Rx_1 \exists x_2 Rx_3 \exists x_4$.

The overall structure of $M(\Phi)$ is sketched in Figure 4. Note that the dashed arrows represent random transitions (all of equal probability, irregardless of the action chosen), solid arrows represent deterministic transitions corresponding to the action True, dotted arrows represent deterministic transitions corresponding to the action False, and dot-dash arrows represent transitions that are forced, whatever the choice of action.

Consider a formula $\Phi \in \text{SSAT}$ with variables $x_1, x_2, \ldots, x_n$ and $r$ random quantifiers, and consider $M(\Phi)$. Because the third stage is deterministic, the process has $2n \cdot 2^r$ trajectories, $n \cdot 2^r$ of which reach stage 3. Now, assume that $\pi$ is a policy, which is *consistent* with the observations from the $n$ steps during the second stage, i.e., whenever it is "asked" to give an assignment to a variable (during the third stage), it does so according to the observations during the second stage and therefore it assigns the same value to every appearance of a variable in $C_{k,a}$. Because $\Phi \in \text{SSAT}$, a fraction of more than $1 - 2^{-c}$ of the trajectories that reach stage 3 correspond to a satisfying assignment and are continued under this policy $\pi$ to state $s_{end}$ where they receive reward 2. Hence, the history-dependent value of $M(\Phi)$ is $> \frac{1}{2} \cdot (1 - 2^{-c}) \cdot 2 = 1 - 2^{-c}$.

---

3. We can also regard the interaction between the process and the policy as an interactive proof system, where the policy presents a proof and the process checks its correctness.

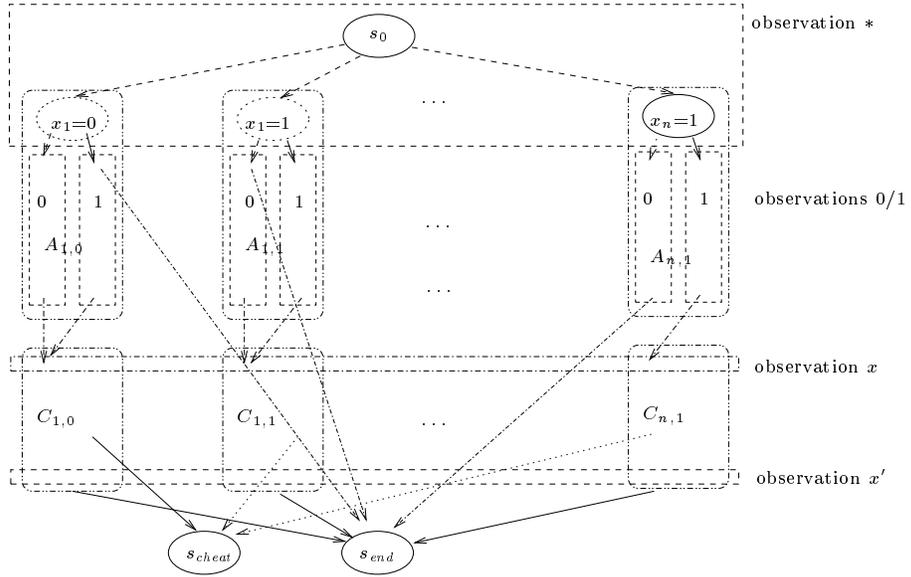

Figure 4: A sketch of $M(\Phi)$.

For $\Phi \notin \text{SSAT}$, an inconsistent (or *cheating*) policy on $M(\Phi)$ may have performance greater than $1 - 2^{-c}$. Therefore, we have to perform a probability amplification as in the proof of Theorem 4.7 that punishes cheating. We construct $M_k(\Phi)$ from $k$ copies $M_1, \ldots, M_k$ of $M(\Phi)$ (the exact value of $k$ will be determined later), such that the initial state of $M_k(\Phi)$ is the initial state of $M_1$, and the initial state of $M_{i+1}$ is the final state $s_{end}$ of $M_i$. If in some repetition a trajectory is caught cheating, then it is sent to the "penalty box" and is not continued in the following repetitions. Hence, it cannot collect any more rewards. More formally, the states $s_{cheat}$ of all the $M_i$s are identified as a unique sink state of $M_k(\Phi)$.

If $\Phi \in \text{SSAT}$, then in each round (or repetition), expected rewards $> 1 - 2^{-c}$ can be collected, and hence the value of $M_k(\Phi)$ is $> k \cdot (1 - 2^{-c})$.

Consider a formula $\Phi \notin \text{SSAT}$. Then a non-cheating policy for $M_k(\Phi)$ has performance less than $k \cdot 2^{-c}$. Cheating policies may have better performances. We claim that for all $k$, the value of $M_k(\Phi)$ is at most $k \cdot 2^{-c} + 2n$. The proof is an induction on $k$. Consider $M_1(\Phi)$, which has the same value as $M(\Phi)$. Hence, the value of $M_1(\Phi)$ is at most 1. As an inductive hypothesis, let us assume that $M_k(\Phi)$ has value at most $k \cdot 2^{-c} + 2n$. In the inductive step, we consider $M_{k+1}(\Phi)$, i.e. $M(\Phi)$ followed by $M_k(\Phi)$. Assume that a policy $\pi_j$ cheats on $j$ of the $2^r$ assignments. From the $n$ trajectories that correspond to an assignment, at least 1 is trapped for cheating under a cheating policy, and at most $n - 1$ may obtain reward 2. Then the reward obtained in the first round is at most $2^{-c} + 2 \cdot \frac{j \cdot (n-1)}{2n \cdot 2^r}$, and the rewards obtained in the following rounds are multiplied by $1 - \frac{j}{2n \cdot 2^r}$, because a fraction of $\frac{j}{2n \cdot 2^r}$ of the trajectories are sent to the penalty box. Using the induction hypothesis, we obtain the following upper bound for the performance of $M_{k+1}(\Phi)$ under $\pi_j$ for an arbitrary $j$.

$$\begin{aligned} \mathit{perf}_f(M_{k+1}(\Phi), \pi_j) &\leq \left(2^{-c} + \frac{j \cdot (n-1)}{n \cdot 2^r}\right) + \left(1 - \frac{j}{2n \cdot 2^r}\right) \cdot \mathrm{val}(M_k(\Phi)) \\ &\leq \left(2^{-c} + \frac{j \cdot (n-1)}{n \cdot 2^r}\right) + \left(1 - \frac{j}{2n \cdot 2^r}\right) \cdot \left(k \cdot 2^{-c} + 2n\right) \\ &= (k+1) \cdot 2^{-c} + 2n - \frac{j}{2^r} \cdot \left(\frac{1}{n} + \frac{k \cdot 2^{-c}}{2n}\right) \\ &\leq (k+1) \cdot 2^{-c} + 2n \end{aligned}$$

This completes the induction step. Hence, we proved that, for $\Phi \notin \textsc{Ssat}$ and for every $k$, the value of $M_k(\Phi)$ is at most $k \cdot 2^{-c} + 2n$.

Eventually, we have to fix the constants. We choose $c$ such that $2^c > \frac{\varepsilon - 2}{\varepsilon - 1}$. This guarantees that
$$(1 - \varepsilon) \cdot (1 - 2^{-c}) - 2^{-c} > 0.$$
Next, we choose $k$ such that
$$2n < k \cdot ((1 - \varepsilon) \cdot (1 - 2^{-c}) - 2^{-c}).$$

Let $\widehat{M(\Phi)}$ be the POMDP that consists of $k$ repetitions of $M(\Phi)$ as described above. Because $k$ is linear in the number, $n$, of variables of $\Phi$ and hence linear in the length of $\Phi$, $\Phi$ can be transformed to $\widehat{M(\Phi)}$ in polynomial time. The above estimates guarantee that

$$\text{value of } \widehat{M(\Phi)} \text{ for } \Phi \notin \textsc{Ssat} \quad \leq \quad k \cdot 2^{-c} + 2n \quad < \quad (1 - \varepsilon) \cdot k \cdot (1 - 2^{-c}).$$

The right-hand side of this inequality is a lower bound for an $\varepsilon$-approximation of the value of $\widehat{M(\Phi)}$ for $\Phi \in \textsc{Ssat}$. Hence,

- if $\Phi \in \textsc{Ssat}$, then $\widehat{M(\Phi)}$ has value $\geq k \cdot (1 - 2^{-c})$, and

- if $\Phi \notin \textsc{Ssat}$, then $\widehat{M(\Phi)}$ has value $< (1 - \varepsilon) \cdot k \cdot (1 - 2^{-c})$.

Hence, a polynomial-time $\varepsilon$-approximation of the value of $\widehat{M(\Phi)}$ shows whether $\Phi$ is in Ssat.

Concluding, let $A$ be any set in PSPACE. There exists a polynomial-time function $f$ which maps every instance $x$ of $A$ to a bounded error stochastic formula $f(x) = \Phi_x$ with error $2^{-c}$ and reduces $A$ to Ssat. Transform $\Phi_x$ into the POMDP $\widehat{M(\Phi_x)}$. Using the $\varepsilon$-approximate value of $\widehat{M(\Phi_x)}$, one can answer "$\Phi_x \in \textsc{Ssat}$?" and hence $x \in A$ in polynomial time. This shows that $A$ is in P, and consequently P = PSPACE. □

**Corollary 4.12** *Let $0 \leq \varepsilon < 1$. The history-dependent value of POMDPs with general rewards is $\varepsilon$-approximable if and only if* P = PSPACE.

## 5. MDPs

Calculating the finite-horizon performance of stationary policies is in GapL (Mundhenk et al., 2000), which is a subclass of the class of polynomial time computable functions. The stationary policy existence problem for MDPs is shown to be P-hard by Papadimitriou and Tsitsiklis (1986), from which it follows that finding an optimal stationary policy for MDPs is P-hard. So it is not surprising that approximating the optimal policy is also P-hard. We include the following theorem because it allows us to present one aspect of the reduction used in the proof of Theorem 5.2 in isolation.

**Theorem 5.1** *The problem of k-additive approximating the optimal stationary policy for MDPs is* P*-hard.*

The proof shows this for the case of non-negative rewards; the unrestricted case follows immediately. By Proposition 3.2, this shows that finding a multiplicative approximation scheme for this problem is also P-hard.

**Proof** Consider the P-complete problem CVP: given a Boolean circuit $C$ and input $x$, is $C(x) = 1$? A Boolean circuit and its input can be seen as a directed acyclic graph. Each node represents a gate, and every gate has one of the types AND, OR, NOT, 0 or 1. The gates of type 0 or 1 are the input gates, which represent the bits of the fixed input $x$ to the circuit. Input gates have indegree 0. All NOT gates have indegree 1, and all AND and OR gates have indegree 2. There is one gate having outdegree 0. This gate is called the output gate, from which the result of the computation of circuit $C$ on input $x$ can be read.

From such a circuit $C$, an MDP $M$ can be constructed as follows. Because the basic idea of the construction is very similar to one shown in (Papadimitriou & Tsitsiklis, 1986), we leave out technical details. As an initial simplifying assumption, assume that the circuit has no NOT gates. Each gate of the circuit becomes a state of the MDP. The start state is the output gate. Reverse all edges of the circuit. Hence, a transition in $M$ leads from a gate in $C$ to one of its predecessors. A transition from an OR gate depends on the action and is deterministic. On action 0 its left predecessor is reached, and on action 1 its right predecessor is reached. A transition from an AND gate is probabilistic and does not depend on the action. With probability $\frac{1}{2}$ the left predecessor is reached, and with probability $\frac{1}{2}$ the right predecessor is reached.

Continue considering a circuit without NOT gates. If an input gate with value 1 is reached, a large positive reward is gained, and if an input gate with value 0 is reached, no reward is gained, which makes the total expected reward noticeably smaller than otherwise. If $C(x) = 1$, then the actions can be chosen at the OR gates so that every trajectory reaches an input gate with value 1; if this condition holds, then it must be that $C(x) = 1$. Hence, the MDP has a large positive value if and only if $C(x) = 1$.

If the circuit has NOT gates, we need to remember the parity of the number of NOT gates on each trajectory. If the parity is even, everything goes as described above. If the parity is odd, then the role of AND and OR gates is switched, and the role of 0 and 1 gates is switched. If a NOT gate is reached, the parity bit is flipped. For every gate in the circuit, we now take two MDP states: one for even and one for odd parity. Hence, if $G$ is the set of gates in $C$, the MDP has states $G \times \{0, 1\}$. The state transition function is

$$t((s,p), a, (s', p')) = \begin{cases} 1, & \text{if } (s \text{ is an OR gate and } p = 0) \text{ or } (s \text{ is an AND gate} \\ & \text{and } p = 1), p = p', \text{ and } s' \text{ is predecessor } a \text{ of } s; \\ \frac{1}{2}, & \text{if } (s \text{ is an OR gate and } p = 1) \text{ or } (s \text{ is an AND gate} \\ & \text{and } p = 0), p = p', \text{ and } s' \text{ is predecessor } 0 \text{ of } s; \\ \frac{1}{2}, & \text{if } (s \text{ is an OR gate and } p = 1) \text{ or } (s \text{ is an AND gate} \\ & \text{and } p = 0), p = p', \text{ and } s' \text{ is predecessor } 1 \text{ of } s; \\ 1, & \text{if } s \text{ is a NOT gate and } s' \text{ is a predecessor of } s \text{ and} \\ & p' = 1 - p; \\ 1, & \text{if } s \text{ is an input gate or the sink state, and } s' \text{ is the sink} \\ & \text{state.} \end{cases}$$

Now we have to specify the reward function. If an input gate with value 1 is encountered on a trajectory where the parity of NOT gates is even, then reward $2^{|C|+k+1}$ is obtained, where $|C|$ is the size of circuit $C$. The same reward is obtained if an input gate with value 0 is encountered on a trajectory where the parity of NOT gates is odd. All other trajectories obtain reward 0.

Thus each trajectory receives reward either 0 or $2^{|C|+k+1}$. There are at most $2^{|C|}$ trajectories for each policy. If a policy chooses the correct values for all the gates in order to prove that $C(x) = 1$, in other words if $C(x) = 1$, then the expected value of an optimal policy is $2^{|C|+k+1}$. Otherwise, the expected value is at least $2^{|C|+k+1}/2^{|C|} \geq 2k$ lower than $2^{|C|+k+1}$, i.e., at most $2^{|C|+k+1} - 2k$.

Thus, if an approximation algorithm is within an additive constant $k$ of the optimal policy, it will either give a value $\geq 2^{|C|+k+1} - k$ or $< 2^{|C|+k+1} - k$. By inspection of the output, one can immediately determine whether $C(x) = 1$. Thus, any $k$-additive approximation for this problem must take at least polynomial time.

In Figure 5, an example circuit and the MDP to which it is transformed are given. Every gate of the circuit is transformed to two states of the MDP: one copy for *even* parity of NOT gates passed on that trajectory (indicated by a thin outline of the state) and one copy for *odd* parity of NOT gates passed on that trajectory (indicated by a thick outline of the state). A solid arrow indicates the outcome of action "choose the left predecessor", and a dashed arrow indicates the outcome of action "choose the right predecessor". Dotted arrows indicate a transition with probability $\frac{1}{2}$ on any action. The circuit in Figure 5 has value 1. The policy, which chooses the right predecessor in the starting state, yields trajectories which all end in an input gate with value 1 and which therefore obtains the optimal value. □

There have been several recent approximation algorithms introduced for structured MDPs, many of which are surveyed in (Boutilier et al., 1999). More recent work includes a variant of policy iteration by Koller and Parr (2000) and heuristic search in the space of finite controllers by Hansen and Feng (2000) and Kim et al. (2000). While these algorithms are often highly effective in reducing the asymptotic complexity and actual run times of policy construction, they all run in time exponential in the size of the structured representation, or offer only weak performance guarantees. We show that exponential asymptotic complexity is necessary for any algorithm scheme that produces $\varepsilon$-approximations for all $\varepsilon$. For this, we consider MDPs represented by 2TBNs (Boutilier, Dearden, & Goldszmidt, 1995). Until now, we have described the state transition function for MDPs by a function $t(s, a, s')$ that computes the probability of reaching state $s'$ from state $s$ under action $a$. We assumed that the transition function was represented explicitly. A *two-phase temporal Bayes net* (2TBN) is a succinct representation of an MDP or POMDP. Each state of the system is described by a vector of values called *fluents*. (Note that if each of $n$ fluents is two-valued, then the system has $2^n$ states.) Actions are described by the effect they have on each fluent by means of two data structures. They are a dependency graph and a set of functions encoded as conditional probability tables, decision trees, arithmetic decision diagrams, or in some other data structure.

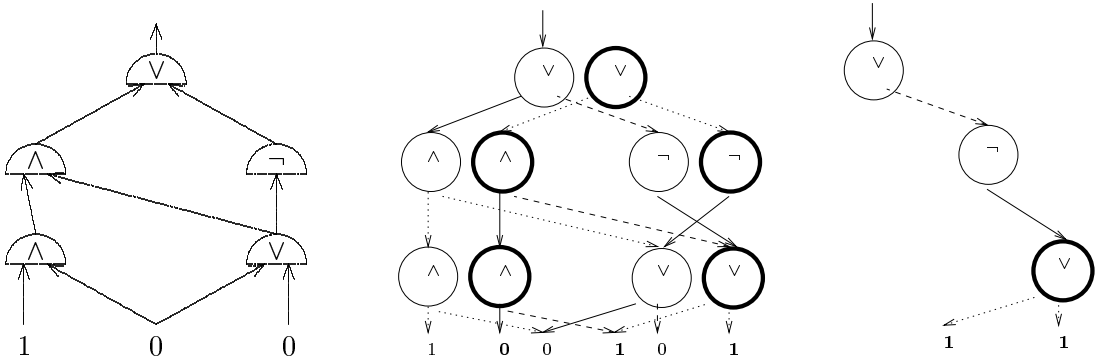

Figure 5: A circuit, the MDP it is reduced to, and the trajectories according to an optimal policy for the MDP

The dependency graph is a directed acyclic graph with nodes partitioned into two sets $\{v_1, \ldots, v_n\}$ and $\{v'_1, \ldots, v'_n\}$. The first set of nodes represents the state at time $t$, the second at time $t+1$. The edges are from the first set of nodes to the second (asynchronous) or within the second set (synchronous). The value of the $k^{th}$ fluent at time $t+1$ under action $a$ depends probabilistically on the values of the predecessors of $v'_k$ in this graph. (Note that the synchronous edges must form a directed, acyclic graph in order for the dependencies to be evaluated.) The probabilities are spelled out, for each action, in the corresponding data structure for $v'_k$ and $a$. We will indicate that (stochastic) function by $f_k$.

We make no assumptions about the structure of rewards for 2TBNs. In fact, the final 2TBN constructed in the proof of Theorem 5.2 has very large rewards which are computed implicitly; in time polynomial in the size of the 2TBN, one can explicitly compute any individual *bit* of the reward. This has the effect of making the potential *value* of the 2TBN too large to write down with polynomially many bits.

**Theorem 5.2** *The problem of k-additive approximating any optimal stationary policy for an MDP in 2TBN-representation is* EXP*-hard.*

**Proof** The general strategy is similar to the proof of Theorem 5.1. We give a reduction from the EXP-complete succinct circuit value problem to the problem for MDPs in 2TBN-representation. An instance of the succinct circuit value problem is a Boolean circuit $S$ that describes a circuit $C$ and an input $x$, i.e. $S$ describes an instance of the "flat" circuit value problem. We can assume that in $C$, each gate is a predecessor to at most two other gates.[4] Then every gate in $C$ has four neighbors, two of which output the input to $C$, and two of which get the output of $C$ as input (if there are fewer neighbors, the missing neighbors are set to a fictitious gate 0). Consider a gate $i$ of $C$. Say that the output of neighbors 0 and 1 is the input to gate $i$, and the output of gate $i$ is input to neighbors 2 and 3. Now, the circuit $S$ on input $(i, k)$ outputs $(j, s)$, where gate $j$ is the $k^{th}$ neighbor of gate $i$, and $s$ encodes the type of gate $i$ (AND, OR, NOT, 0, and 1). The idea is to construct from $C$ an MDP $M$ as in the proof of Theorem 5.1. However, we do it succinctly. Hence, we construct from $S$ a 2TBN-representation of an MDP $M(S)$. The actions of $M(S)$ are 0 and 1, for choosing neighbor 0 of the current state-gate, or respectively, neighbor 1. The states of $M(S)$ are tuples $(i, p, t, r)$ where $i$ is a gate of $C$, $p$ is the parity bit — as in the proof of Theorem 5.1, $t$ is the type of gate $i$, and $r$ is used for a random bit. Every gate number $i$ is given in binary using — say — $l$ bits. Then, the 2TBN has $l + 3$ fluents $i_1, i_2, \ldots, i_l, p, t, r$. Let $f_1, f_2, \ldots, f_l, f_p, f_t, f_r$ be the stochastic functions that calculate $i'_1, i'_2, \ldots, i'_l, p', t', r'$. The simplest is $f_r$ for the fluent $r'$ that is used as random bit if from state $i = i_1 \cdots i_l$ the next state is chosen uniformly at random from one of the two predecessors of gate $i$ in $C$. This happens if the type $t$ of gate $i$ is AND and the parity $p$ is 0, or if $t$ is OR and $p$ is 1. In these cases, $r'$ determines its value 0 or 1 by flipping a coin. Otherwise, $r'$ equals 1. Notice that $r'$ is independent of the action.

The functions $f_c$ for the fluents $v'_c$ determine the bits of the next states. If $t$ is an AND and the parity $p$ is even, then "randomly" a predecessor of gate $i$ is chosen. "Randomly" means here that the random bit $r'$ determines whether predecessor 0 or predecessor 1 is chosen. Hence, $v'_c$ is the $c^{th}$ bit of $j$, where $(j, s)$ is the output of $S$ on input $(i, r')$. Accordingly, $t' = s$ is the type of the chosen gate, and $p' = p$ remains unchanged. The same happens if $t$ is an OR and the parity $p$ is odd. If $t$ is a NOT, there is only one predecessor of $i$, and that one must be chosen for $i'$ and $t'$. The parity bit $p'$ is flipped to $1 - p$. If $t$ is an OR and the parity $p$ is even, then on action $a \in \{0, 1\}$, the predecessor $a$ of gate $i$ is chosen. Hence, $v'_c$ is the $c^{th}$ bit of $j$, where $(j, s)$ is the output of $S$ on input $(i, a)$. Accordingly, $t' = s$ and $p' = p$. The same happens if $t$ is an AND and the parity $p$ is odd. Hence, the function $f_c$ can be calculated as follows.

> **input** $i, p, t, r', a$
> **if** ($t = $ OR and $p = 0$) or ($t = $ AND and $p = 1$)
>   **then** calculate $S(i, a) = (j, s)$;
> **else if** ($t = $ OR and $p = 1$) or ($t = $ AND and $p = 0$)
>   **then** calculate $S(i, r') = (j, s)$
> **else if** $t = $ NOT
>   **then** calculate $S(i, 0) = (j, s)$
> **else** $j = \mathbf{0}$
> **output** the $c^{th}$ bit of $j$

The state $\mathbf{0}$ is a sink state which is reached from the input gates within one step and which is never left. The type $t'$ of the next state or gate is calculated accordingly.

---

4. If this is not the case, and $d$ is the maximum out-degree of a gate, we can replace the circuit by one with maximum out-degree 2 and size at most $\log d$ larger. Since $d \leq |C|$, such a substitution will not affect the asymptotic complexity of any of our algorithms.

Lusena, Goldsmith, & Mundhenk

One can also simulate the circuit $S$ for function $f_k$ in the above algorithm by a 2TBN. Note that, in general, circuits can have more than one output. We consider this more general model here.

**Claim 1** *Every Boolean circuit can be simulated by a 2TBN, to which it can be transformed in polynomial time.*

**Proof** We sketch the construction idea. Let $R$ be a circuit with $n$ input gates and $n'$ output gates. The outcome of the circuit on any input $b_1, \ldots, b_n$ is usually calculated as follows. At first, calculate the outcome of all gates that get input only from input gates. Next, calculate the outcome of all gates that get their inputs only from those gates whose outcome is already calculated, and so on. This yields an enumeration of the gates of a circuit in topological order, i.e., such that the outcome of a gate can be calculated when all the outcomes of gates with a smaller index are already calculated. We assume that the gates are enumerated in this way, and that $g_1, \ldots, g_n$ are the input gates, and that $g_l, \ldots, g_s$ are the other gates, where the smallest index of a gate which is neither an output nor an input gate equals $l = \max(n, n') + 1$.

Now, we define a 2TBN $T$ simulating $R$ as follows. $T$ has a fluent for every gate of $R$, say fluents $v_1, \ldots, v_s$. The basic idea is that fluents $v_1, \ldots, v_n$ represent the input gates of $R$. In one time step, values are propagated from the input nodes $v_1, \ldots, v_n$ through all gate nodes $v'_l, \ldots, v'_s$, and the outputs copied to $v'_1, \ldots, v'_{n'}$. The dependency graph has the following edges according to the "wires" of the circuit $R$. If an input gate $g_i$ ($1 \leq i \leq n$) outputs an input to gate $g_j$, then we get an edge from $v_i$ to $v'_j$. If the output of a non-input gate $g_i$ ($n < i < s$) is input to gate $g_j$, then we get an edge from $v'_i$ to $v'_j$. Finally, the nodes $v'_1, \ldots, v'_{n'}$ stand for the value bits. If gate $g_j$ produces the $i$th output bit, then there is an edge from $v'_j$ to $v'_i$. Because the circuit $R$ has no loop, the graph is loop-free, too.

The functions associated to the nodes $v'_1, \ldots, v'_s$ depend on the functions calculated by the respective gate and are as follows. Each of the value nodes $v'_i$ for $i = 1, 2, \ldots, n'$, which stands for the input bits, has exactly one predecessor, whose value is copied into $v'_i$. Hence, $f_i$ is the one-place identity function, $f_i(x) = x$ with probability 1, for $i = 1, 2, \ldots, n'$. Now we consider the nodes which come from internal gates of the circuit. If $g_i$ is an AND gate, then $f_i(x, y) = x \wedge y$, where $x$ and $y$ are the predecessors of gate $g_i$. If $g_i$ is an OR gate, then $f_i(x, y) = x \vee y$, and if $g_i$ is a NOT gate, then $f_i(x) = \neg x$, all with probability 1.

By this construction, it follows that the 2TBN $T$ simulates the Boolean circuit $R$. Notice that the number of fluents of $T$ is at most the double of the number of gates of $R$. The transformation from $R$ to $T$ can be performed in polynomial time. □

An example of a Boolean circuit and the 2TBN to which it is transformed as described above is given in Figure 6.

Now, we can construct from the circuit $S$ that is a succinct representation of a circuit $C$ a 2TBN $T_S$ with fluents $i_1, i_2, \ldots, i_l, p, t, r$ as already defined, plus additional fluents for the gates of $S$, using the technique from the above Claim. Taking the action $a$, the parity $p$, the gate type $t$ and the random bit $r'$ into account, we can construct $T_S$ — according to the description of function $f_c$ above — so that fluent $v'_{j_c}$ contains the bit described by the function $f_c(i_1, i_2, \ldots, i_l, p, t, r)$ above. Notice that the function $f_{j_c}$ for $v'_{j_c}$ is dependent only on the predecessors of the gate of $S$ represented by $v'_{j_c}$, the fluents $p, t, r'$, and the action $a$.

106

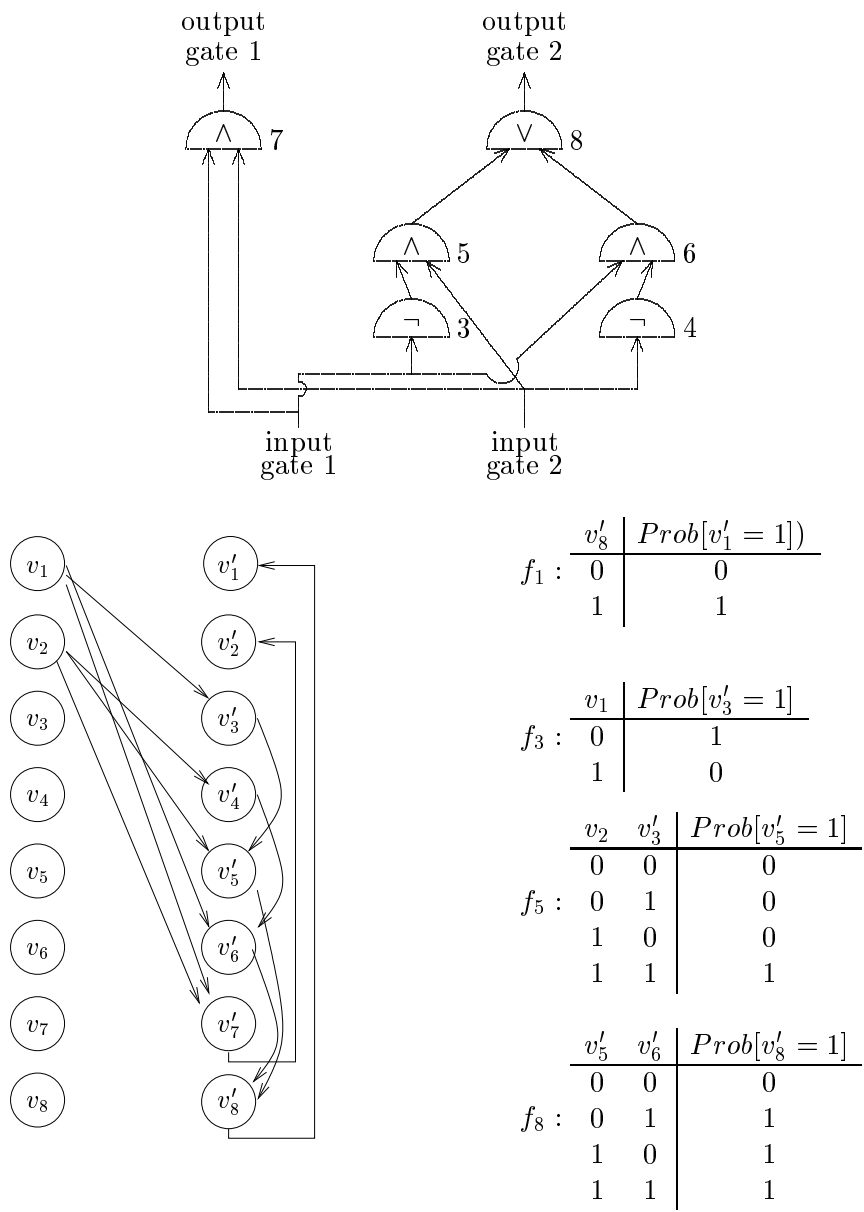

Figure 6: A Boolean circuit which outputs the binary sum of its input bits, and a 2TBN representing the circuit. Only functions $f_1$ (the identity function), $f_3$ (simulating a NOT gate), $f_5$ (simulating an AND gate), and $f_8$ (simulating an OR gate) are described).

Hence, it has at most 6 arguments and can be described by a small table. This holds for all fluents of $T_S$. Finally, the function $f_c$ for $T_S$ just copies the value of $v'_{j_c}$ into $v'_c$. Hence, from $S$ we can construct a MDP in 2TBN representation similar to the MDP in the proof of Theorem 5.1. Next, we specify the rewards of this MDP. The reward is $2^{2^{|S|+k+1}}$ if any action is taken on a state representing an input gate with value 1 and parity 0, or with value 0 and parity 1. Otherwise, the reward equals 0. This reward function can be represented by a circuit, which on binary input $i, a, b$ outputs the $b^{th}$ bit of the reward obtained in state $i$ on action $a$. (Since it requires $2^{|S|+k+1}$ bits to represent the reward, $b$ can be represented using only $|S| + k + 1$ bits.)

If $C(x) = 1$, then there is a choice of actions for each state that gives reward $2^{2^{|S|+k+1}}$ on every trajectory, similar to the proof of Theorem 5.1. However, if $C(x) = 0$, any policy has at least one trajectory that receives a 0 reward. Now, there are at most $2^{2^{|S|}}$ trajectories, and therefore there is a gap of at least $2^{2^{|S|+k+1}}/2^{2^{|S|}} \geq 2k$ between possible values. As above, we conclude that any $k$-additive approximation to the factored MDP problem gives a decision algorithm for the succinct circuit value problem. Therefore, the lower bound of EXP-hardness for the factored MDP value problem holds for this approximation problem as well. □

The following structured representation is more general than the representations more common to the AI/planning community. We say that an MDP has a *succinct representation*, or is a succinct MDP, if there are Boolean circuits $C_t$ and $C_r$ such that $C_t(s, a, s', i)$ produces the $i$th bit of the transition probability $t(s, a, s')$ and $C_r(s, a, i)$ produces the $i$th bit of the reward $r(s, a)$. Similar to the proof of Theorem 5.2, we can also prove nonapproximability of MDP values for succinctly represented MDPs.

**Theorem 5.3** *The problem of $k$-additive approximating the optimal stationary policy for a succinctly represented MDP is* EXP*-hard.*

## 6. Non-Approximability for Infinite-Horizon POMDPs

The discounted value of an infinite-horizon POMDP is the maximum total discounted performance. When we discuss the policy existence problem or the average case performance in the infinite horizon, it is necessary to specify the reward criterion. We generalize the value function as follows.

The $\alpha, \beta$-*value* $\text{val}_{\alpha,\beta}(M)$ of $M$ is $M$'s maximal $\beta$-performance under any policy $\pi$ of type $\alpha$, i.e. $\text{val}_{\alpha,\beta}(M) = \max_{\pi \in \Pi_\alpha} \text{perf}_\beta(M, \pi)$.

Note that a time-dependent or history-dependent infinite-horizon policy for a POMDP is not necessarily finitely representable. For fully-observable MDPs, it turned out (see e.g. Puterman, 1994) that the discounted or average value is the performance of a stationary policy. This means that no history-dependent policy performs better than the best stationary one. As an important consequence, an optimal policy is finitely representable. For POMDPs, this does not hold. Madani et al. (1999) showed that the time-dependent infinite-horizon policy-existence problem for POMDPs is not decidable under average performance or under total discounted performance. In contrast, we show that the same problem for stationary policies is NP-complete.

**Theorem 6.1** *The stationary infinite-horizon policy-existence problem for POMDPs under total discounted or average performance is* NP*-complete.*

The hardness proof is essentially the same as for Theorem 4.1. Note that in that construction, every stationary policy obtains reward 1 for at most one step, namely when sink state $T$ is reached, meaning that the formula is satisfied. All other steps yield reward 0. Therefore, for this construction, the total discounted value is greater than 0 if and only if the finite-horizon value is so. To make the construction work for average value, we have to modify it such that once the sink state $T$ is reached, every subsequent action brings reward 1. Therefore, the average value equals 1 if the formula is satisfiable, and it equals 0 if it is unsatisfiable. Hence, both the problems are NP-hard.

Containment in NP for the total discounted performance follows from the guess-and-check approach: Guess a stationary policy, calculate its performance and accept if and only if the performance is positive. The total discounted and the average performance can both be calculated in polynomial time.

In the same way, the techniques proving nonapproximability results for the stationary policy in the finite horizon case (Corollary 4.2) can be modified to obtain nonapproximability results for infinite horizons.

**Theorem 6.2** *The stationary infinite-horizon value of POMDPs under total discounted or average performance can be $\varepsilon$-approximated if and only if* P = NP.

The infinite-horizon time-dependent policy-existence problems are undecidable (Madani et al., 1999). We show that no computable function can even approximate optimal policies.

**Theorem 6.3** *The time-dependent infinite-horizon value of unobservable POMDPs under average performance cannot be $\varepsilon$-approximated.*

The proof follows from the proof by Madani et al. (1999) showing the uncomputability of the time-dependent value. In Madani et al. (1999), from a given Turing machine $T$ an unobservable POMDP is constructed having the following properties for arbitrary $\delta > 0$. (1) If $T$ halts on empty input, then there is exactly one time-dependent infinite-horizon policy with performance $\geq 1 - \delta$, (2) all other time-dependent policies have performance $\leq \delta$, and (3) the average value is between 0 and 1. This reduces the undecidable problem of whether a Turing machine halts on empty input to the time-dependent infinite-horizon policy existence problem for unobservable POMDPs under average performance. Actually, assuming that the value of the unobservable POMDP were $\varepsilon$-approximable, we could choose $\delta$ in a way that even the approximation enables us to decide whether $T$ halts on empty input. Since this is undecidable, an $\varepsilon$-approximation is impossible.

**Corollary 6.4** *The time-dependent and history-dependent infinite-horizon value of POMDPs under average performance cannot be $\varepsilon$-approximated.*

### Acknowledgements

This work was supported in part by NSF grant CCR-9315354 and CCR-9610348 (Lusena and Goldsmith), and by Deutscher Akademischer Austauschdienst (DAAD) grant 315/PPP/gü-ab (Mundhenk). The third author performed part of the work at Dartmouth College.

We would like to thank Daphne Koller and several anonymous referees for catching errors in earlier versions of this paper.

## Appendix A. Proof of Theorem 4.1

We present the reduction from (Mundhenk et al., 2000). Let $\phi(x_1, \ldots, x_n)$ be an instance of 3SAT with variables $x_1, \ldots, x_n$ and clauses $C_1, \ldots, C_m$, where clause $C_j = (l_{v(1,j)} \vee l_{v(2,j)} \vee l_{v(3,j)})$ for $l_i \in \{x_i, \neg x_i\}$. We say that variable $x_i$ *appears in $C_j$ with signum 0 (resp. 1)* if $\neg x_i$ (resp. $x_i$) is a literal in $C_j$.

From $\phi$, we construct a POMDP $M(\phi) = (\mathcal{S}, s_0, \mathcal{A}, \mathcal{O}, t, o, r)$ with

$$\begin{aligned}
\mathcal{S} &= \{(i,j) \mid 1 \leq i \leq n, 1 \leq j \leq m\} \cup \{F, T\} \\
s_0 &= (v(1,1), 1), \quad \mathcal{A} = \{0,1\}, \quad \mathcal{O} = \{x_1, \ldots, x_n, F, T\}
\end{aligned}$$

$$t(s, a, s') = \begin{cases} 1, & \text{if } s = (v(i,j), j), s' = (v(1, j+1), j+1), j < m, 1 \leq i \leq 3, \\ & \quad \text{and } x_{v(i,j)} \text{ appears in } C_j \text{ with signum } a \\ 1, & \text{if } s = (v(i,m), m), s' = T, 1 \leq i \leq 3, \\ & \quad \text{and } x_{v(i,m)} \text{ appears in } C_m \text{ with signum } a \\ 1, & \text{if } s = (v(i,j), j), s' = (v(i+1, j), j), 1 \leq i < 3, \\ & \quad \text{and } x_{v(i,j)} \text{ appears in } C_j \text{ with signum } 1-a \\ 1, & \text{if } s = (v(3,j), j), s' = F, \\ & \quad \text{and } x_{v(3,j)} \text{ appears in } C_j \text{ with signum } 1-a \\ 1, & \text{if } s = s' = F \text{ or } s = s' = T \\ 0, & \text{otherwise} \end{cases}$$

$$r(s,a) = \begin{cases} 1, & \text{if } t(s, a, T) = 1, s \neq T \\ 0, & \text{otherwise} \end{cases}, \quad o(s) = \begin{cases} x_i, & \text{if } s = (i,j) \\ T, & \text{if } s = T \\ F, & \text{if } s = F \end{cases}.$$

Note that all transitions in $M(\phi)$ are deterministic, and every trajectory has value 0 or 1. There is a correspondence between policies for $M(\phi)$ and assignments of values to the variables of $\phi$, such that policies under which $M(\phi)$ has value 1 correspond to satisfying assignments for $\phi$, and vice versa.

## Appendix B. Proof of Theorem 4.6

Again, we present the reduction from (Mundhenk et al., 2000). Let $\phi$ be a formula with $n$ variables $x_1, \ldots, x_n$ and $m$ clauses $C_1, \ldots, C_m$. This time, we define the unobservable MDP $M(\phi) = (\mathcal{S}, s_0, \mathcal{A}, t, r)$ where

$$\begin{aligned}
\mathcal{S} &= \{(i,j) \mid 1 \leq i \leq n, 1 \leq j \leq m\} \cup \{sat_i \mid 1 \leq i \leq n\} \cup \{s_0, T, F\} \\
\mathcal{A} &= \{0, 1\}
\end{aligned}$$

$$t(s,a,s') = \begin{cases} \frac{1}{m}, & \text{if } s = s_0, s' = (1,j), 1 \leq j \leq m \\ 1, & \text{if } s = (i,j), s' = sat_{i+1}, i < n, \ x_i \text{ appears in } C_j \text{ with signum } a \\ 1, & \text{if } s = (i,j), s' = (i+1,j), i < n, \\ & \quad x_i \text{ does not appear in } C_j \text{ with signum } a \\ 1, & \text{if } s = (n,j), s' = T, \ x_n \text{ appears in } C_j \text{ with signum } a \\ 1, & \text{if } s = (n,j), s' = F, \ x_n \text{ does not appear in } C_j \text{ with signum } a \\ 1, & \text{if } s = sat_i, s' = sat_{i+1}, i < n \\ 1, & \text{if } s = sat_n, s' = T \\ 1, & \text{if } s = s' = F \text{ or } s = s' = T, a = 0 \text{ or } a = 1 \\ 0, & \text{otherwise} \end{cases}$$

$$r(s,a) = \begin{cases} m, & \text{if } s \neq T \text{ and } t(s,a,T) > 0 \\ 0, & \text{otherwise} \ . \end{cases}$$